\documentclass[letterpaper, 10 pt, conference]{ieeeconf}  
\usepackage{times}  
\usepackage{helvet}  
\usepackage{courier}  
\usepackage[hyphens]{url}  
\usepackage{graphicx} 
\usepackage{caption} 
\usepackage{pifont}
\usepackage{amsmath}
\usepackage{amsfonts}
\usepackage[graphicx]{realboxes}
\usepackage{graphicx}
\usepackage{textcomp}
\usepackage{booktabs}
\usepackage{fancyhdr}
\usepackage{multirow}
\usepackage{color}
\usepackage{bm}
\usepackage{enumerate} 
\usepackage{mathrsfs}
\usepackage{makeidx}           
\usepackage{multicol}      
\usepackage{psfrag}

\usepackage[ruled,vlined]{algorithm2e}
\usepackage{algorithmic}
\usepackage{colortbl}  
\usepackage[dvipsnames]{xcolor}
\usepackage{array} 
\usepackage{makecell}
\usepackage{algorithmic}
\usepackage{amsmath}
\usepackage{color}
\usepackage{newfloat}
\usepackage{listings}
\usepackage{hhline}

\IEEEoverridecommandlockouts                              

\title{\LARGE \bf
Fed3D: Federated 3D Object Detection
}

\author{Suyan Dai$^{1}$, Chenxi Liu$^{2,*}$, Fazeng Li$^{1}$, Peican Lin$^{1}$
\\
\thanks{$^{1}$School of Automation Science and Enginnering, South China University of Technology, Guangzhou 510641, P.R.China.}
\thanks{$^{2}$State Key Laboratory of Robotics, Shenyang Institute of Automation,
Chinese Academy of Sciences, Shenyang, 110016, China; Institutes for Robotics and Intelligent Manufacturing, Chinese Academy
of Sciences, Shenyang, 110169, China.}
\thanks{$^{*}$The corresponding author is Mr. Chenxi Liu.}
}

\begin{document}

\maketitle
\thispagestyle{empty}
\pagestyle{empty}

\begin{abstract}
3D object detection models trained in one server plays an important role in autonomous driving, robotics manipulation, and augmented reality scenarios. However, most existing methods face severe privacy concern when deployed on a multi-robot perception network to explore large-scale 3D scene. Meanwhile, it is highly challenging to employ conventional federated learning methods on 3D object detection scenes, due to the 3D data heterogeneity and limited communication bandwidth. In this paper, we take the first attempt to propose a novel \underline{Fed}erated \underline{3D} object detection framework (\emph{i.e.,} Fed3D), to enable distributed learning for 3D object detection with privacy preservation. Specifically, considering the irregular input 3D object in local robot and various category distribution between robots could cause local heterogeneity and global heterogeneity, respectively. We then propose a local-global class-aware loss for the 3D data heterogeneity issue, which could balance gradient back-propagation rate of different 3D categories from local and global aspects. To reduce communication cost on each round, we develop a federated 3D prompt module, which could only learn and communicate the prompts with few learnable parameters. To the end, several extensive experiments on federated 3D object detection show that our Fed3D model significantly outperforms state-of-the-art algorithms with lower communication cost when providing the limited local training data.

\end{abstract}
\definecolor{deepred}{rgb}{0.698,0.133,0.133}

\section{INTRODUCTION}

3D object detection \cite{d:8, zhou2018voxelnet} is developed to localize and categorize 3D object from a 3D dataset as a efficient technique to understand 3D scene. Recently, this technique has achieved great advancement in many real-word fields, such as autonomous driving \cite{chen2017multi}, robotics manipulation \cite{cong2021comprehensive}, augmented reality \cite{rukhovich2022fcaf3d} and so on. Compared with 2D object detection scenario, 3D data are irregular and spare due to different sensors. Therefore, most existing 3D object detection methods are proposed via extracting point-wise or voxel-wise semantic features, and predicting a set of proposals that will be fed to the detection head. For example, VoteNet \cite{d:1} builds a relationship between the 3D semantic features and 3D object center, and votes a series of proposals; \cite{zhang2020h3dnet} further improves the point
group generation procedure. However, these methods assume that 3D training data is easy to access without privacy risks, and aim to train a 3D detector by gathering data from one learner. When extended into a multi-robot 3D perception network in distributed learning scenarios, \emph{e.g.,} autonomous driving or collaborative robots, these methods have to learn new knowledge from robots while accumulating the knowledge in the server under the limitation of privacy protection. This great challenge motivates us to adopt the 3D object detection tasks into a federated machine learning setting.       
\begin{figure}[t]
	\centering
	\includegraphics[width=230pt,height=160pt]
	{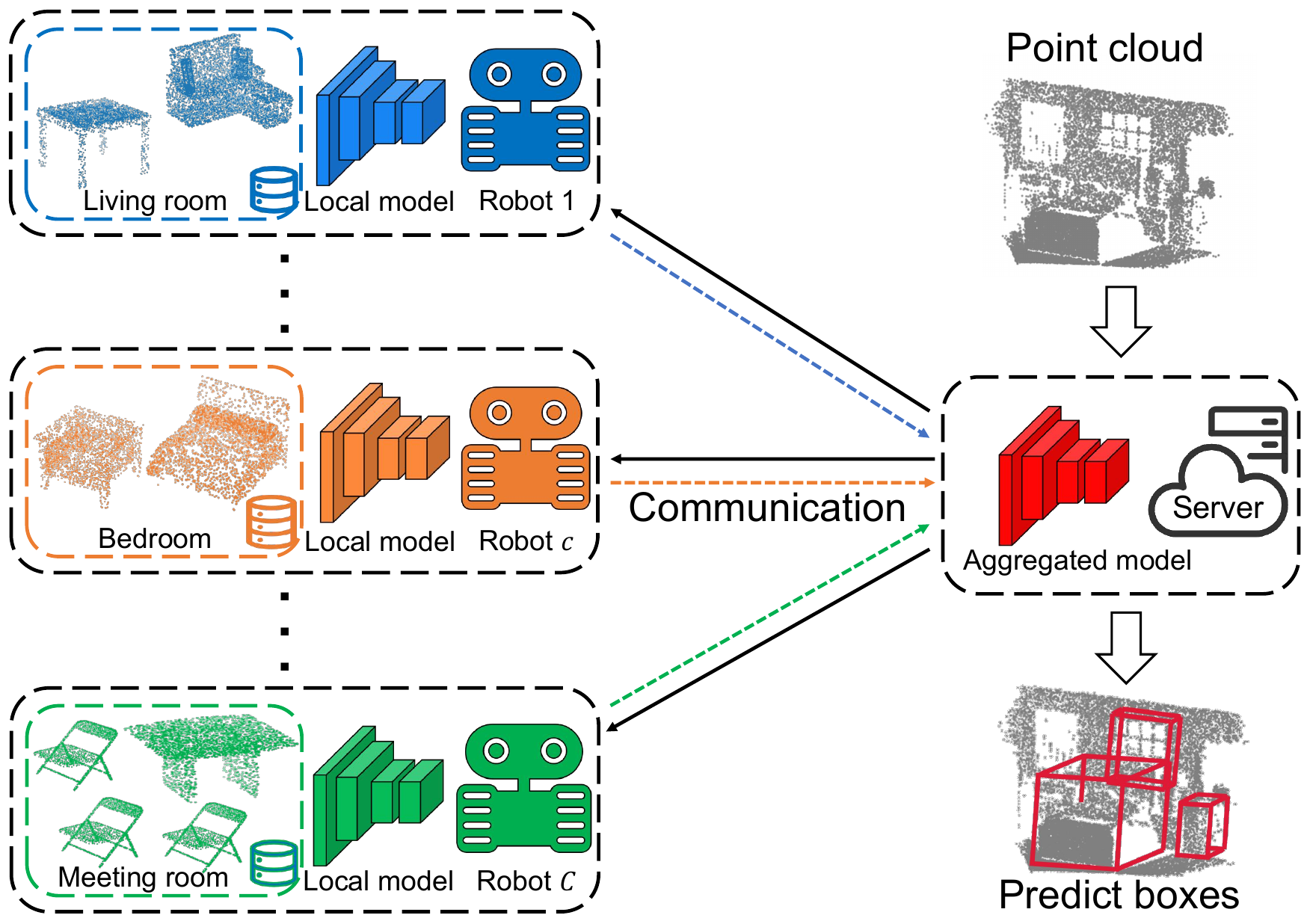}
 	\vspace{-5pt}
	\caption{Illustration of our proposed Fed3D model to address the distributed robotic perception issue.} 
	\label{fig: motivation}
	\vspace{-15pt}
\end{figure}

Federated learning (FL) \cite{thapa2022splitfed, liu2021feddg, yang2021federated} is proposed to train a powerful global model on a server with the help of a series of local clients without sharing local private data. Specifically, federated leaning aggregates model weights of models of clients after performing multiple local training, thereby protecting the privacy of clients. For instance, FedAvg  \cite{f:2} is developed to enhance local training to reduce communication costs between global server and local clients. However, two challenges arise when federated learning is applied to 3D object perception scenarios: \textbf{3D data heterogeneity} and \textbf{limited communication bandwidth}. For example, as shown in multi-robot system in Fig.~\ref{fig: motivation}, local heterogeneity arises when irregular and imbalanced 3D data in robot (\emph{e.g.}, in a meeting room scene, the point cloud of chair has different object shape and quantity compared with the point cloud of table). In addition, the global imbalance of category will lead to global heterogeneity (\emph{e.g.}, the categories in a bedroom scene are different from those in a living room scene when two robots are distributed in bedroom and living room). Directly adapting existing federated 2D object detection methods to perform on the 3D scenes will ignore local and global heterogeneity. Furthermore, these methods suffer from high communication cost, which is difficultly applied to edge device with limited communication bandwidth, such as mobile robots and vehicles.

To tackle the above problems, we propose the first \underline{Fed}erated \underline{3D} object detection framework, called as Fed3D, which can achieve multi-robot collaborative 3D object detection with limited communication requirements. Specifically, to tackle \textbf{3D data heterogeneity} in federated 3D perception scenarios, we introduce a local-global class-aware loss to tackle 3D data heterogeneity from local and global aspects. We find that both local and global 3D data imbalance issues will lead to local and global 3D data heterogeneity in Fed3D. For this reason, we develop a a simple yet effective technique to measure local and global imbalance. Then we introduce a re-weighted loss function to control gradient back-propagation speed of each sample and mitigate the impact on the update of model parameters caused by local and global imbalance. As for the \textbf{limited communication bandwidth}, we develop a prompt learning module in our Fed3D model. To our knowledge, we are the first attempt to develop prompt tuning in training a 3D detector. Specifically, we build a local prompt pool contains a few parameters to store 3D knowledge learned by local robot, and only upload or download prompt pool to perform model aggregation in each global communication. Meanwhile, only a few parameters are transmitted between global server and local robots, which can significantly reduce the communication cost. Furthermore, the model attackers and global server can only obtain a prompt pool rather than the whole model from local robots, which can efficiently avoid privacy leaks. 

When applied to a system with edge device to perform federated 3D object detection tasks, our Fed3D model can train a global 3D detector to detect all 3D objects seen by all robots without invading the privacy of robots. For each communication round, Fed3D selects a subset of all robots to perform balanced local prompt learning with a local-global class-aware loss. Then the global server collects local prompt pools, and aggregates them to obtain a global prompt pool that is distributed to all robots for the next communication round. To the end, we conduct expensive FL experiments on ScanNet V2 \cite{dai2017scannet} and SUN RGB-D \cite{song2015sun} datasets. Experimental results suggest that our Fed3D model achieves a better performance in comparison to the state-of-the-art methods with a smaller communication cost. We summarize our contributions as follows:

\begin{itemize}
\item We propose a novel federated 3D object detection framework, Fed3D, for the first time, which could efficiently perform multi-robot (\emph{e.g.}, robots and vehicles) collaborative 3D scene perception without destroying privacy of local robots.
\item  A local-global class-aware loss is designed to address 3D data heterogeneity issue from gradient aspect, which can mitigate the impact of local data imbalance in local robots and global 3D category imbalance on server. 
\item A prompt learning strategy is introduced to perform federated 3D object detection, which could only transmit a few parameters in each communication round, thereby reducing the communication cost and protecting privacy of local clients. Expensive experiments verify the significant performance of our method compared with the state-of-the-arts.  
\end{itemize}

\begin{figure*}[t]
	\centering
	\includegraphics[width=490pt, height=245pt]
	{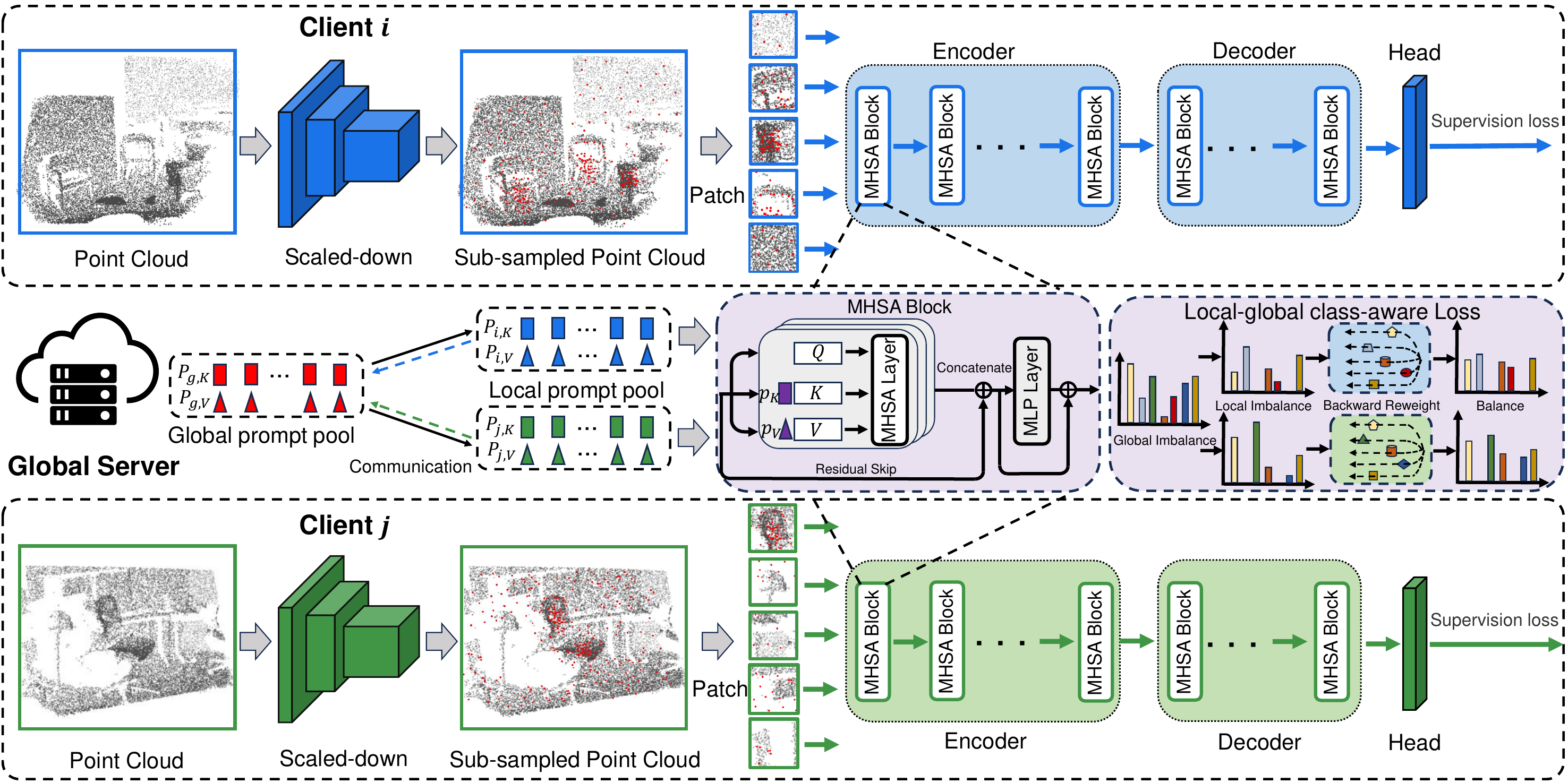}
 \caption{Overview framework of our method Fed3D, whose network structure  is based on PiMAE \cite{chen2023pimae}. It mainly consists of a federated prompt learning to reduce communication cost between global server and local clients, a local-global class-aware loss to mitigate the impact on the update of model parameters caused by local and global imbalance.} 
	\label{fig: overview_of_our_model}
	\vspace{-10pt}
\end{figure*}

\section{RELATED WORK}
\subsection{Federated Learning}
Federated learning \cite{f:2,f:1}, as a distributed machine learning paradigm, has garnered substantial attention for its ability to effectively preserve user data privacy and security \cite{f:12,f:10,f:14,f:15}. FedAvg \cite{f:2} first proposes training a global model by simply averaging client model weights without transmitting data, which limits performance due to non-IID data. To address this, FedProx \cite{f:3} enhances model performance through re-parameterization of FedAvg. Scaffold \cite{f:4} curbs client drift during local updates by regulating variables. Moon \cite{f:5} rectifies client bias during local training by harnessing similarity between model representations.  FedDyn and FedX \cite{f:8,f:10} take recourse to distillation method.  FedAlign \cite{f:16} leverages an efficacious regularization approach. From the security perspective, Sun \emph{et al.}~\cite{sun2022data} investigated data poisoning attacks on federated machine learning and showed that the communication protocol among clients can be exploited to derive effective poisoning strategies. Specifically, they formulated the poisoning process in federated multi-task learning as a bilevel optimization problem and proposed a systems-aware attack method, namely AT2FL. This line of work reveals that keeping training data local does not necessarily guarantee the robustness and security of federated learning systems. For 2D object detection, FedVision \cite{f:6} enables city-scale video surveillance through federated computer vision. However, these methods discount local and global imbalance of 3D point clouds, and incur high communication costs when deployed on edge devices.


\subsection{3D Objection Detection}
3D  data harbors abundant depth information and has been extensively employed in domains including autonomous driving and robot navigation. To effectively acquire 3D scene context, numerous CNN \cite{d:4,d:9,d:1,d:3,d:11,d:10} and Transformer-based \cite{d:8,d:5,d:6,d:7} networks are proposed. VoteNet \cite{d:1} first introduces rough voting on PointNet++ \cite{d:2} features for 3D data processing to estimate offsets and cluster object centers.  Due to the great success of Transformers in other computer vision tasks, 3DETR \cite{d:5} spearheads an end-to-end 3D detection framework based on Transformers. GroupFree3D \cite{d:6} supplants conventional voting with Transformers and achieves better performance. Despite this,  the aforementioned 3D detectors often exhibit catastrophic failure in the face of non-IID data structure and could not be extended into a multi-robot 3D perception network in distributed learning scenarios.

\section{METHOD}
\subsection{Problem Definition and Overview}

For federated 3D object detection, we follow the experimental setting in Federated Learning (FL)~\cite{f:2}. Assume that there are $C$ local clients, and $c$-th client contains a 3D detector $\omega_c$ and a local 3D  dataset $\mathcal{D}_c$. Due to local and global  heterogeneity in Fed3D, $P_{m\in\mathcal{D}_c} \not\sim P_{m\in\mathcal{D}_j}$ and $P_{s\in\mathcal{D}_c} \not\sim P_{s\in\mathcal{D}_j}$, where $m$ represents 3D category and $s$ denotes 3D sample. Fed3D aims to train a global 3D detector on the dataset $\cup_{c=1}^{C}\mathcal{D}_c$ with a distributed learning manner. The training objective function for federated 3D object detection can be formulated as:
\vspace{-2pt}
\begin{align}
       \label{eq: eq1}
       \mathop{\arg\min}\limits_\omega \mathcal{F}(\omega) = \sum_{c=1}^C \frac{\lvert\mathcal{D}_c\rvert}{\lvert\mathcal{D}\rvert}\mathcal{F}_c (\omega_c), 
\end{align}\vspace{-10pt}
\begin{align}
       \label{eq: eq2}
       \mathcal{F}_c(\omega_c) = \frac{1}{\lvert\mathcal{D}_c\rvert} \sum_{i=1}^{\mathcal{D}_c}\mathcal{L}^c_i (\omega_c;x^c_i,y^c_i), 
\end{align}
where $\lvert\mathcal{D}\rvert$ denotes the quantity of samples in $\mathcal{D}$, and $\mathcal{F}_c(\omega_c)$ is the local objective of client $c$, and $\mathcal{L}^c_i (\omega_c;x^c_i,y^c_i)$ represents the loss function of model $\omega_c$ on a pair of 3D data and target $(x^c_i,y^c_i)$. Specifically, by following FedAvg \cite{f:2}, we select randomly a subset of all clients with a certain proportion $\alpha$ to start local training on the local datasets on each communication round. Then the global server collects the local model parameters of selected clients and aggregate them to obtain global model as $\omega_g=\sum_{c=1}^S \frac{\lvert\mathcal{D}_c\rvert}{\lvert\mathcal{D}\rvert}\omega_c$, where $S$ denotes the number of selected clients.

The overview framework of our Fed3D is shown in Fig.~\ref{fig: overview_of_our_model}. Following PiMAE \cite{chen2023pimae}, we feed the original point cloud into a scale-down network to obtain the sub-sampled point cloud.  Then the patch features will be sent to our pre-training ViT encoder where we introduce prompt learning.  A local-global class-aware loss is introduced to our supervision loss, which could measure data imbalance from local and global aspects, and tackle local and global imbalance from gradient back propagation. Meanwhile,  we build a local prompt pool to store the prompt pairs that are inserted into layers of encoder. For each communication round, global server aggregates local prompt pools uploaded by selected clients to obtain a global prompt pool and distribute it to all clients.

\subsection{Local-global Class-aware Loss}
Catastrophic 3D data imbalance will mislead deep neural network to a wrong optimization direction (\emph{i.e.}, overfitting) and result in notorious data heterogeneity in FL. Most of exiting methods tackle data imbalance through data augmentation strategy or re-weighting loss. However, these methods cannot be applied to FL, due to the fact that their violation of the privacy requirements and ignoring the global imbalance in FL. In federated 3D object detection, we define two kinds of 3D data imbalance: local and global imbalance. Specifically, local imbalance is caused by the variation of the quantity and irregular shape of 3D data for different classes. The class distribution varies among local clients, which leads global imbalance in FL.

To address the two aspects of data imbalance, we introduce a local-global class-aware observer to measure 3D data imbalance. Specifically, inspired by \cite{wang2021addressing}, we observe the impact of local and global imbalance from gradient aspect. In each mini-batch training, the update of local model parameters can be simplified as:
\begin{align}
       \label{eq: eq3}
       \mathcal{\omega}^{n+1} \xleftarrow{} \mathcal{\omega}^{n}-\frac{\eta}{B}\sum^O_{o=1}\sum^{B_o}_{i=1}\nabla_{\mathcal{\omega}^n_{i,o}}\mathcal{L} (\mathcal{\omega}^{n};x_i,y_i), 
\end{align}
where $B$ represents the quantity of samples in a mini-batch and $B_o$ is the quantity of samples of class $o$. $O$ denotes the total number of 3D classes and $\mathcal{\omega}^{n}$ is the model parameters in $n$-th batch. According to \cite{wang2021addressing}, if the similar output logits $\hat{y}_{i,o}$ are induced from the samples of a same class, the gradients of these samples are also very similar. However, the output logits of different samples vary due to the data imbalance. Thus we can revise this variation to mitigate the impact on the update of local model parameters caused by data imbalance in Eq.~\ref{eq: eq3}. Specifically, we follow \cite{dong2022federated} to measure the gradient with respect to the $\hat{y}_i$-th neuron $\mathcal{N}_{\hat{y}_i}$ of the last output layer in $\omega$ by :
\begin{align}
       \label{eq: eq4}
       \nabla_{\mathcal{N}_{\hat{y}_i}}\mathcal{L}_{CE} (\mathcal{\omega};x_i,y_i) = \lvert\sigma(\hat{y}_i)-1\rvert,
\end{align}
where $\sigma$ denotes the softmax operation, and $\mathcal{L}_{CE}$ represents cross entropy loss for classification. The value of gradient in Eq.~\ref{eq: eq4} could reflect and measure influence on back propagation process brought by data imbalance. In light of this, we measure 3D data imbalance from local and global aspects.
\renewcommand{\algorithmicrequire}{\textbf{Input:}}
\renewcommand{\algorithmicensure}{\textbf{Output:}}

\begin{algorithm}[t]			
    	\caption{Optimization pipeline of our Fed3D.} 
	\label{alg: optimization}
	\begin{algorithmic}[1]
		\REQUIRE The local clients $\{\mathcal{D}_c\}_{c=1}^C$, communication round $Z$, local training epoch $E$;
  \STATE \textbf{\#Server}:	
  \STATE Initialize global prompt pool $\gamma^1_g$, model $\hat{\omega}^1_g$ and global correction coefficient $\cup^O_{o=1}\mathcal{G}^o$.
  \FOR {$z=1, 2, \cdots, Z$}

        \STATE  Randomly select participants $\{\mathcal{D}_c\}_{c=1}^S$;
        \FOR {Each $\mathcal{D}_c$ $\in$ $\{\mathcal{D}_c\}_{c=1}^S$}
          \STATE $\gamma^{z,E}_c, \hat{\omega}^{z,E}_c,\mathcal{R}^{c,o}_g  \xleftarrow{}$ Local($\cup^O_{o=1}\mathcal{G}^o; \gamma^z_g,\hat{\omega}^z_g$);

                         \ENDFOR 
         \STATE Update $\gamma^{z+1}_g,\hat{\omega}^{z+1}_g$ by Eq.~\ref{eq: eq15};
        \STATE Update $\cup^O_{o=1}\mathcal{G}^o$ by Eq.~\ref{eq: eq7};  
		\ENDFOR \\
  \STATE return $\gamma^{Z}_g,\hat{\omega}^{Z}_g$
    \STATE \textbf{\#Local}($\cup^O_{o=1}\mathcal{G}^o; \gamma^z_g,\hat{\omega}^z_g$):
         \STATE  Initialize $\gamma^{z,1}_c,\hat{\omega}^{z,1}_c \xleftarrow{} \gamma^z_g,\hat{\omega}^z_g$;
     \FOR {$e=1, 2, \cdots, E$}
            \STATE Update $\gamma^{z,e+1}_c,\hat{\omega}^{z,e+1}_c$ by Eq.~\ref{eq: eq14};     

  \ENDFOR 

 \STATE return $\gamma^{z,E}_c,\hat{\omega}^{z,E}_c$   
\end{algorithmic} 

\end{algorithm}
  \begin{figure*}[t]
	\centering
	\includegraphics[width=500pt, height=165pt]
	{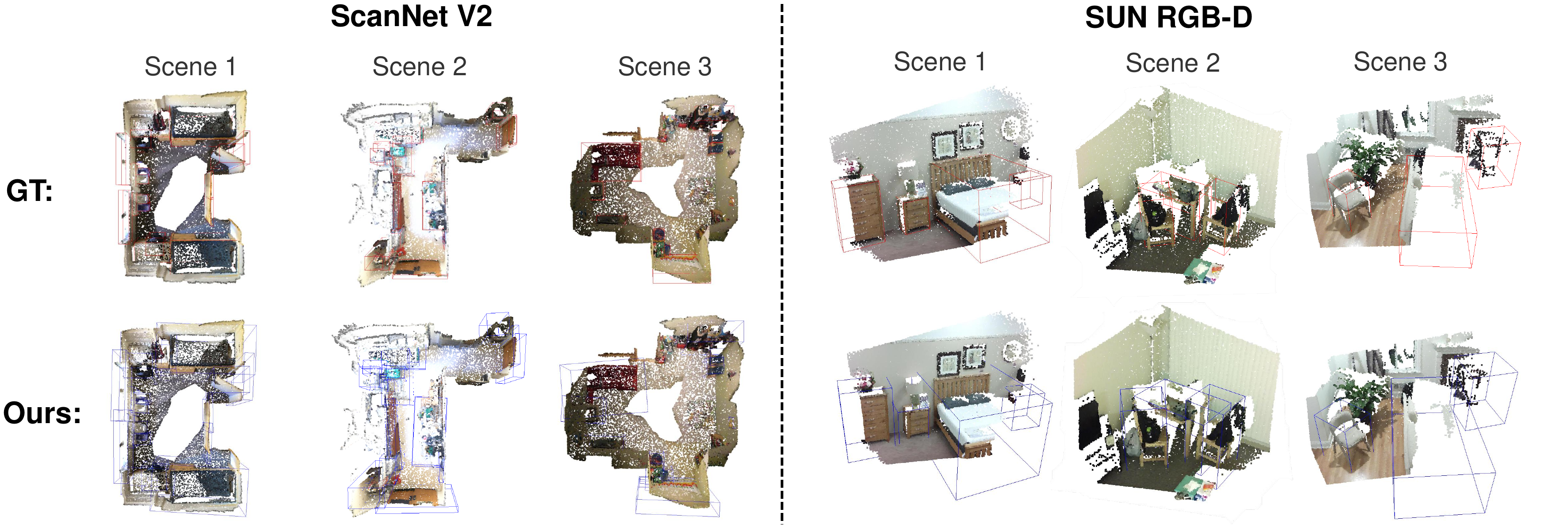}
 	\vspace{-5pt}
	\caption{Visualization results of ground truth (GT) and predict of ours on two benchmark datasets. The left row is the results on ScanNet V2~\cite{dai2017scannet}, and the right row is the results on SUN RGB-D~\cite{song2015sun}.} 
	\label{fig: visual}
	\vspace{-15pt}
\end{figure*}
Specifically, we utilize the measurement in Eq.~\ref{eq: eq4} to estimate local imbalance, and obtain a  correction coefficient to re-weight back propagation in Eq.~\ref{eq: eq3}. The local correction coefficient in the batch training is computed as follows:
\begin{align}
       \label{eq: eq5}
\mathcal{R}_i=\frac{B\nabla_{\mathcal{N}_{\hat{y}_i}}\mathcal{L}_{CE} (\mathcal{\omega};x_i,y_i)} {\sum_{b=1}^B \nabla_{\mathcal{N}_{\hat{y}_b}}\mathcal{L}_{CE} (\mathcal{\omega};x_b,y_b)}.
\end{align}

Furthermore, for global imbalance, we need to estimate 3D class imbalance but cannot directly measure as Eq.~\ref{eq: eq5} due to lacking 3D data in global server. Thus we will measure global imbalance though local clients without violating their privacy. In this way, we define the representation of class distribution for class $o$ in $c$-th client as follows:
\begin{align}
       \label{eq: eq6}
\mathcal{R}^{c,o}_g=\frac{1}{\sum_{n=1}^{N_c} \mathbb{I}_{y_n=o}}\sum_{n=1}^{N_c}\nabla_{\mathcal{N}_{\hat{y}_n}}\mathcal{L}_{CE} (\mathcal{\omega}_c;x^c_n,y^c_n)  \cdot \mathbb{I}_{y_n=o},
\end{align}
where $N_c$ denotes the number of samples in $c$-th client and $\mathbb{I}$ means the indicator function that $\mathbb{I}=1$ when the condition is true; or $\mathbb{I}=0$. $\mathcal{R}^{c,o}_g$ suggests the imbalance among various classes, and we introduce it to Eq.~\ref{eq: eq5} to update the $o$-th global correction coefficient as follows:
\begin{align}
       \label{eq: eq7}
\mathcal{G}^o =\mathrm{ln}(1+2\frac{SO\sum_{c=1}^S\mathcal{R}^{c,o}_g}{\sum_{c=1}^S\sum_{o=1}^O\mathcal{R}^{c,o}_g}).
\end{align}

Then we apply $\mathcal{G}^o$ to Eq.~\ref{eq: eq4}, and replace $\nabla_{\mathcal{N}_{\hat{y}_i}}\mathcal{L}_{CE} (\mathcal{\omega};x_i,y_i)$ with $\nabla_{\mathcal{N}_{\hat{y}_i}}\mathcal{L}_{CE} (\mathcal{\omega};x_i,y_i)^{\sum_{o=1}^O\mathcal{G}^o\cdot \mathbb{I}_{y_i=o}}$ to obtain  $\hat{\mathcal{R}_i}$ to perform the local-global correction.
The update of local model parameters in Eq.~\ref{eq: eq3}  during $n$-th mini batch training can be revised as:
\begin{align}
       \label{eq: eq8}
       \mathcal{\omega}^{n+1} \xleftarrow{} \mathcal{\omega}^{n}-\frac{\eta}{B}\sum^O_{o=1}\sum^{B_o}_{i=1}\hat{\mathcal{R}_i}\nabla_{\mathcal{\omega}^n_{i,o}}\mathcal{L} (\mathcal{\omega}^{n};x_i,y_i). 
\end{align}

To this end, we propose a local-global class-aware loss defined in Eq.~\ref{eq: eq8} to revise the update of local model parameters 
 and  mitigate the impact caused by local-global imbalance, which can efficiently address 3D data heterogeneity. 
\subsection{Federated Prompt Learning for 3D object detection}
Federated learning suffers the high communication bandwidth when applied to multi-robots system. To address this challenge, we introduce  prompt tuning to train a 3D detector for the first time. Specifically, for the learning of backbone, we build a local prompt pool     $\gamma_c=\{P^1_c, \cdot\cdot\cdot, P^l_c, \cdot\cdot\cdot, P^L_c \}$, where $L$ denotes the number of layers in encoder and $P^l_c$ represents the prompt embed in $l-$th layer. For each communication round, we update local prompt pool during local training. Then we aggregate all local prompt pools from selected clients to  obtain a global prompt pool, and distribute global prompt pool to all clients to revise their local prompts pools. In light of this, only a few parameters rather than the whole model weights are communicated between global server and local clients, thereby significantly addressing the high communication cost. 
As shown in Fig.~\ref{fig: overview_of_our_model}, we introduce a self-supervised pre-trained encoder from PiMAE \cite{chen2023pimae} as the backbone of our 3D detector. Given a 3D point cloud $x$, we feed it to a scaled-down PointNet \cite{d:2} to extra a series of sub-sampled points $\hat{x}$ and low-level point cloud feature $f$. Then we apply our frozen encoder and local prompt pool to handle $f$. Concretely, based on transformer, encoder consists of multi-head self-attention (MHSA) blocks, where feedforward process of MHSA block can be simplified as:
\begin{align}
	\label{eq: eq9}
\mathrm{A}^{l,h} =  \sigma(\frac{Q^{l,h} \mathrm{W}_Q^{l,h} (K^{l,h} \mathrm{W}_K^{l,h})^{\top}}{\sqrt{d}})(V^{l,h}  \mathrm{W}_V^{l,h}),
\end{align}  
\begin{align}
	\label{eq: eq10}
\mathrm{O}^l =  f^{l-1} +\mathrm{Concat}(\mathrm{A}^{l,1},\cdot\cdot\cdot,\mathrm{A}^{l,H})\mathrm{W}^l_O,
\end{align}  
\begin{align}
	\label{eq: eq11}
f^{l}= f^{l-1} + \mathrm{MLP}(\mathrm{O}^l), 
\end{align}  
where $\mathrm{W}^l_O, \mathrm{W}_Q^{l,h},\mathrm{W}_K^{l,h},\mathrm{W}_V^{l,h}$ are mapping matrices, and $\sigma$ represents softmax function, and $\sqrt{d}$ is the channel dimension of each head. In addition, $f = \mathrm{Concat}(Q^{l,1},\cdot\cdot\cdot, Q^{l,H})$ and $H$ denotes the number of heads, where $Q^{l,h}=K^{l,h}=V^{l,h}$. Inspired by Prefix-tuning \cite{li2021prefix}, we extend Eq.~\ref{eq: eq9} to a cross-attention manner using a pair of prefix prompts as follows:
\begin{align}
	\label{eq: eq12}
\mathrm{A}^{l,h} =  \sigma(\frac{\hat{Q}^{l,h} \mathrm{W}_Q^{l,h} (\hat{K}^{l,h} \mathrm{W}_K^{l,h})^{\top}}{\sqrt{d}})(\hat{V}^{l,h} \mathrm{W}_V^{l,h}),
\end{align}  
where $\hat{K}^{l,h}=[K^{l,h}, P^{l,h}_K], \hat{V}^{l,h}=[V^{l,h}, P^{l,h}_V]$, where $P^l= \mathrm{Concat}[(P^{l,1}_K,P^{l,1}_V),\cdot\cdot\cdot,(P^{l,H}_K,P^{l,H}_V)]$. Compared with universal prompt tuning, prefix prompt tuning could remain the same sequence length between the output feature and input feature. To this end, we combine all $P^l$ of each layer to build a local prompt pool, and learn the backbone of our 3D detector through updating the prompt pool. The local objective of client $c$ can be updated as follows:
\begin{align}
       \label{eq: eq13}
       \mathcal{F}_c(\gamma_c,\hat{\omega}_c) = \frac{1}{\lvert\mathcal{D}_c\rvert} \sum_{i=1}^{\mathcal{D}_c}\mathcal{L}^c_i (\gamma_c,\hat{\omega}_c;x^c_i,y^c_i), 
\end{align}
where $\hat{\omega}_c$ denotes other weights in 3D detector $\omega_c$ except backbone. Compared with the whole model weights $\omega_c$, $(\gamma_c, \hat{\omega}_c)$ only accounts for a small part of the whole parameters, which could efficiently reduce the communication bandwidth.
Furthermore, the global server and attackers that intercept the communication between global server and local clients could not restore the whole local model from the prompt pools, thereby preserving the privacy of local clients.
\begin{table*}[t]
\centering
\setlength{\tabcolsep}{0.80mm}
\renewcommand\arraystretch{1.0}
\large
\caption{Comparison results on ScanNet V2~\cite{dai2017scannet} dataset in terms of mAP and AR.}
\vspace{-5pt}
\scalebox{0.75}{
\begin{tabular}{c|c|cc|cc|cc|cc|c|cc|cc|cc|cc} \hline
	\toprule
	\multirow{3}{*}{Methods} & \multicolumn{9}{c|}{$C$=20, $\alpha$=0.25}  & \multicolumn{9}{c}{$C$=100, $\alpha$=0.1} \\\cline{2-19} &\makecell[c]{\multirow{2}{*}{\#Com.cost}}
 & \multicolumn{4}{c|}{IoU=0.25} & \multicolumn{4}{c|}{IoU=0.5} &\makecell[c]{\multirow{2}{*}{\#Com.cost}} & \multicolumn{4}{c|}{IoU=0.25}   & \multicolumn{4}{c}{IoU=0.5} \\ 
	& & mAP & Imp.   & AR & Imp. & mAP & Imp.   & AR & Imp. & & mAP & Imp.   & AR & Imp. & mAP & Imp.   & AR & Imp. \\ 
	\midrule
	Local-training &86.2M  &19.1  &$\Uparrow$27.0 & 50.2 &$\Uparrow$32.8 &10.9  & $\Uparrow$16.0 & 46.2 &$\Uparrow$6.8 &86.2M   &19.1  &$\Uparrow$28.4 & 50.2 &$\Uparrow$32.8 &10.9  & $\Uparrow$19.3 & 46.2 &$\Uparrow$9.5    \\
	Centralized &86.2M  &60.3  & $\Downarrow$14.2 & 86.7 &$\Downarrow$3.7  &35.8  & $\Downarrow$5.9 & 73.2 &$\Downarrow$23.2 &86.2M   &60.3  & $\Downarrow$12.8 & 86.7 &$\Downarrow$3.3 &35.8  & $\Downarrow$5.6 & 73.2 &$\Downarrow$17.5    \\
 \midrule
 	FedAvg & 86.2M & 41.3 &$\Uparrow$4.8  & 81.6 & $\Uparrow$1.4& 25.8 & $\Uparrow$1.1  & 52.0 & $\Uparrow$1.0 & 86.2M  &43.4 & $\Uparrow$4.1  &81.0  & $\Uparrow$2.4 & 29.0 & $\Uparrow$1.2 &52.9 &$\Uparrow$2.8 \\
  	FedProx  &86.2M  & 42.3 & $\Uparrow$3.8 & 81.5 &$\Uparrow$1.5 & 26.1  &$\Uparrow$0.8 & 51.9  & $\Uparrow$1.1  &86.2M &45.8 &$\Uparrow$1.7  &81.3  &$\Uparrow$2.1  & 28.9 &$\Uparrow$1.3  &53.8 &$\Uparrow$1.5  \\
    SCAFFOLD  &86.2M  &42.2  &$\Uparrow$3.9  & 81.6 &$\Uparrow$1.4 &24.7  &$\Uparrow$2.2  & 48.9 &$\Uparrow$4.1  &86.2M   &45.2 & $\Uparrow$2.3 & 80.8 &$\Uparrow$2.6  &29.6  & $\Uparrow$0.6 &55.3 &$\Uparrow$0.4  \\
    MOON &86.2M  &45.0  &$\Uparrow$1.1  & 82.0 &$\Uparrow$1.0 & 26.5 &$\Uparrow$0.4   & 52.2 &$\Uparrow$0.8  &86.2M   &46.3 & $\Uparrow$1.2  &83.0  &$\Uparrow$0.4  &29.3  & $\Uparrow$0.9 &54.8 &$\Uparrow$0.9 \\
    FedDyn  &86.2M  &43.3  &$\Uparrow$2.8  & 81.4 &$\Uparrow$1.6 & 22.3 &$\Uparrow$4.1  & 49.0&$\Uparrow$4.0  & 86.2M  &46.7 & $\Uparrow$0.8  &81.6  &$\Uparrow$1.8  &30.0  & $\Uparrow$0.2 &55.0 &$\Uparrow$0.7  \\
    Ours  &\textcolor{deepred}{\textbf{43.6M}}  &\textcolor{deepred}{\textbf{46.1}}  & $\mathrm{-}$   &\textcolor{deepred}{\textbf{83.0}}  &$\mathrm{-}$ &\textcolor{deepred}{\textbf{26.9}} &$\mathrm{-}$ &\textcolor{deepred}{\textbf{53.0}} &$\mathrm{-}$  &\textcolor{deepred}{\textbf{43.6M}}   &\textcolor{deepred}{\textbf{47.5}}  & $\mathrm{-}$  &\textcolor{deepred}{\textbf{83.4}} &$\mathrm{-}$  &\textcolor{deepred}{\textbf{30.2}} &$\mathrm{-}$  &\textcolor{deepred}{\textbf{55.7}}  &$\mathrm{-}$  \\
	
	\bottomrule \hline
\end{tabular}}
\label{tab: scanet}  
\vspace{-5pt}
\end{table*}

\begin{table*}[ht]
\centering
\setlength{\tabcolsep}{0.80mm}
\renewcommand\arraystretch{1.0}
\large
\caption{Comparison results on SUN RGB-D~\cite{song2015sun} dataset in terms of mAP and AR.}
\vspace{-5pt}
\scalebox{0.75}{
\begin{tabular}{c|c|cc|cc|cc|cc|c|cc|cc|cc|cc} \hline
	\toprule
	\multirow{3}{*}{Methods} & \multicolumn{9}{c|}{$C$=20, $\alpha$=0.25}  & \multicolumn{9}{c}{$C$=100, $\alpha$=0.1} \\\cline{2-19} &\makecell[c]{\multirow{2}{*}{\#Com.cost}}
 & \multicolumn{4}{c|}{IoU=0.25} & \multicolumn{4}{c|}{IoU=0.5} &\makecell[c]{\multirow{2}{*}{\#Com.cost}} & \multicolumn{4}{c|}{IoU=0.25}   & \multicolumn{4}{c}{IoU=0.5} \\ 
	& & mAP & Imp.   & AR & Imp. & mAP & Imp.   & AR & Imp. & & mAP & Imp.   & AR & Imp. & mAP & Imp.   & AR & Imp. \\ 
	\midrule
	Local-training &86.2M  &12.3  &$\Uparrow$30.6 & 48.3 &$\Uparrow$33.5 &8.4  & $\Uparrow$16.4 & 32.2 &$\Uparrow$17.7 &86.2M   &12.3  &$\Uparrow$31.7 & 48.3 &$\Uparrow$33.6 &8.4  & $\Uparrow$16.6 & 32.2 &$\Uparrow$22.1    \\
	Centralized &86.2M  &58.3  & $\Downarrow$15.4 & 83.8 &$\Downarrow$2.0  &33.0  & $\Downarrow$8.2 & 71.6 &$\Downarrow$21.7 &86.2M   &58.3  & $\Downarrow$14.3 & 83.8 &$\Downarrow$1.9  &33.0  & $\Downarrow$8.0 & 71.6 &$\Downarrow$17.3    \\
 \midrule
 	FedAvg & 86.2M & 39.0 &$\Uparrow$3.9  & 79.8 & $\Uparrow$2.0 & 20.4 & $\Uparrow$4.4  & 48.3 & $\Uparrow$1.6 & 86.2M  &40.1 & $\Uparrow$3.9  &80.1  & $\Uparrow$1.8 & 20.1 & $\Uparrow$4.9 &50.0 &$\Uparrow$4.3 \\
  	FedProx  &86.2M  & 39.9 & $\Uparrow$3.0 &80.3 &$\Uparrow$1.5 & 23.3  &$\Uparrow$1.6 & 50.2  & $\Downarrow$0.3  &86.2M &40.8 &$\Uparrow$3.2  &81.6  &$\Uparrow$0.3  & 21.4 &$\Uparrow$3.6  &50.7 &$\Uparrow$3.7  \\
    SCAFFOLD  &86.2M  &40.7  &$\Uparrow$2.1  & 80.0 &$\Uparrow$1.8 &22.3  &$\Uparrow$2.5  & 45.6 &$\Uparrow$4.3  &86.2M   &40.9 & $\Uparrow$3.1 & 78.9 &$\Uparrow$3.0  &23.5  & $\Uparrow$1.5 &54.0 &$\Uparrow$0.3  \\
    MOON &86.2M  &42.1  &$\Uparrow$0.8  & 80.1 &$\Uparrow$1.7 & 22.2 &$\Uparrow$2.6   & 47.9 &$\Uparrow$2.0  &86.2M   &41.1 & $\Uparrow$2.9  &79.8  &$\Uparrow$2.1  &22.0  & $\Uparrow$3.0 &51.2 &$\Uparrow$3.1 \\
    FedDyn  &86.2M  &41.9  &$\Uparrow$1.0  & 81.7 &$\Uparrow$0.1 & 24.4 &$\Uparrow$0.4  &\textcolor{deepred}{\textbf{51.1}} &$\Downarrow$1.2  & 86.2M  &43.8 & $\Uparrow$0.2  &80.9  &$\Uparrow$1.0  &24.4  & $\Uparrow$0.6 &52.2 &$\Uparrow$2.1  \\
    Ours  &\textcolor{deepred}{\textbf{43.6M}}  &\textcolor{deepred}{\textbf{42.9}}  &$\mathrm{-}$   &\textcolor{deepred}{\textbf{81.8}}  &$\mathrm{-}$ &\textcolor{deepred}{\textbf{24.8}} &$\mathrm{-}$ &49.9 &$\mathrm{-}$  &\textcolor{deepred}{\textbf{43.6M}}   &\textcolor{deepred}{\textbf{44.0}}  & $\mathrm{-}$  &\textcolor{deepred}{\textbf{81.9}} &$\mathrm{-}$  &\textcolor{deepred}{\textbf{25.0}} &$\mathrm{-}$  &\textcolor{deepred}{\textbf{54.3}}  &$\mathrm{-}$  \\
	
	\bottomrule \hline
\end{tabular}}
\label{tab: sunrgbd}  
\vspace{-10pt}
\end{table*}
\label{tab: scannet-class}  

\begin{table}[ht]
\centering
\setlength{\tabcolsep}{0.62mm}
\renewcommand\arraystretch{1.1}
\large
\caption{Ablation studies on ScanNet V2~\cite{dai2017scannet} dataset in terms of mAP and AR.}
\vspace{-5pt}
\scalebox{0.59}{
\begin{tabular}{c|c|cc|cc|cc|cc} \hline
	\toprule
	\multirow{3}{*}{Methods} & \multicolumn{9}{c}{$C$=20, $\alpha$=0.25}  \\\cline{2-10} &\makecell[c]{\multirow{2}{*}{\#Com.cost}}
 & \multicolumn{4}{c|}{IoU=0.25} & \multicolumn{4}{c}{IoU=0.5} \\
	 & & mAP & Imp.   & AR & Imp. & mAP & Imp.   & AR & Imp. \\ 
	\midrule
	Local-training & 86.2M & 19.1 &$\Uparrow$27.0  & 50.2 & $\Uparrow$32.8 &10.9 &$\Uparrow$16.0  &46.2 &$\Uparrow$6.8     \\
	Centralized & 86.2M & 60.3 &$\Downarrow$14.2  & 86.7 & $\Downarrow$3.7 &35.8 &$\Downarrow$5.9  &73.2 &$\Downarrow$23.2    \\
 \midrule
 	Baseline &86.2M & 41.3 &$\Uparrow$4.8  & 81.6 & $\Uparrow$1.4 & 25.8 & $\Uparrow$1.1  & 52.0 & $\Uparrow$1.0   \\
  	Ours w/o FPL \& GCC  &86.2M  &44.8  &$\Uparrow$1.3  &82.0 &$\Uparrow$1.0 & 26.0 &$\Uparrow$0.2 & 52.4 &$\Uparrow$0.6    \\
    Ours w/o FPL \& LCC  &86.2M  &42.8  &$\Uparrow$3.3  &81.2 &$\Uparrow$1.8 & 26.2 &$\Uparrow$0.7 & 52.3 &$\Uparrow$0.7     \\
     Ours w/o FPL &86.2M  &45.8  &$\Uparrow$0.3  &82.5 &$\Uparrow$0.5 & 26.7 &$\Uparrow$0.2 & 52.6 &$\Uparrow$0.4    \\
    Ours  &\textcolor{deepred}{\textbf{43.6M}}  &\textcolor{deepred}{\textbf{46.1}} &$\mathrm{-}$ &\textcolor{deepred}{\textbf{83.0}}   &$\mathrm{-}$ &\textcolor{deepred}{\textbf{26.9}}  &$\mathrm{-}$ &\textcolor{deepred}{\textbf{53.0}}  &$\mathrm{-}$ \\
	
	\bottomrule \hline
\end{tabular}}
\label{tab: ablation}  
\vspace{-10pt}
\end{table}

\vspace{-10pt}
\textbf{Local Update Step}: In each communication round $z =
\{1, 2, ..., Z\}$, the clients are trained using the following update rules with a learning rate of $\eta_c$:
\begin{align}
       \label{eq: eq14}
       \gamma^{t+1}_c, \hat{\omega}^{t+1}_c \xleftarrow{} \gamma^{t}_c, \hat{\omega}^{t}_c-\frac{\eta_c}{\lvert\mathcal{D}_c\rvert} \sum_{i=1}^{\mathcal{D}_c}\hat{\mathcal{R}_i}\nabla_{\hat{\mathcal{\omega}}^t_{i}}\mathcal{L} (\gamma_c^t,\hat{\mathcal{\omega}}_c^{t};x_i,y_i),
\end{align}
where $t$ denotes the $t$-th update of the local clients.

\textbf{Server Update Step}: After a round of local updates, all
participating clients send their updated prompts to the
server performing aggregation. Such process can be expressed as follows:
\begin{align}
       \label{eq: eq15}
(\gamma^{z+1}_g, \hat{\omega}^{z+1}_g) \xleftarrow{} \sum_{c=1}^C \frac{\lvert\mathcal{D}_c\rvert}{\lvert\mathcal{D}\rvert}(\gamma^z_c, \hat{\omega}^z_c).
\end{align}
To sum up, the issue can be largely mitigated because
there are only a small number of learnable parameters 
communicated between the server and local clients. After $Z$
rounds of communication, we can get a robust global model
parameterized by $\gamma_g^t$ without sharing local private data.

\section{EXPERIMENTS}

\subsection{Datasets and Evaluation}
We conduct experiments on two widely-used indoor 3D object detection datasets: SUN RGB-D \cite{song2015sun} and ScanNet V2 \cite{dai2017scannet} and we prepare them  following VoteNet \cite{d:1}.

\textbf{SUN RGB-D} is a large-scale RGB-D dataset with rich 3D indoor objects, consisting of 10,335 RGB-D images and depth images for 37 object categories. Following VoteNet, we perform the standard 5,285, 5,050 splits with the 10 selected  catagories for the training and test datasets.  

\textbf{ScanNet V2} is a 3D indoor scene dataset with  enriched annotations, consisting
of 1,513 indoor scenes and 18 object categories. 
For federated 3D object detection, the standard mean Average Precision (mAP) and Average Recall (AR) in terms of IoU thresholds of 0.25 and 0.5 are
utilized to valuate the performance between Ours and compared methods.

\subsection{Experimental Setup}

\textbf{Implementation Details}.
Following PiMAE \cite{chen2023pimae}, we build a 3D detector for all experiments utilizing a scaled-down PointNet \cite{d:2}, a 6 layers ViT encoder pre-trained on SUN RGB-D in a self-supervised manner, a 8 layers ViT decoder and a series of detection head. For federated 3D object detection, we develop two federated client settings (\emph{i.e.}, $C=100, \alpha=0.1$ and $C=20, \alpha=0.25$), where $C$ denotes the number of clients and $\alpha$ represents the selection ratio of clients for each communication round. To simulate 3D class heterogeneity and sample heterogeneity, we distribute 70\% classes with 70\% samples for each client on two benchmark datasets.  We conduct all experiments with the 100 global communication rounds and the 4 local training epochs. 
As for the local training, we utilize 50k input points with a batch size 8, and adopt the same data augmentation as in PiMAE, including a random flip, a random rotation, and
a random scaling of the point cloud. In addition, we train local 3D detector utilizing an AdamW optimizer with a learning rate $3.5 \times 10^4$ for encoder and  $7.0 \times 10^4$ for other modules. 

\textbf{Baselines}. To valuate the effectiveness of our Fed3D, we apply two methods with different training manner and five standard federated  methods to perform federated 3D object detection: 1) Local-training, a single local model that each client trained on their local data; 2) Centralized, a single global model trained on the whole data; In addition, five Fl methods are adopted under Fed3D settings: 3) FedAvg \cite{f:2}, 4) FedProx \cite{f:3}, 5) SCAFFOLD \cite{f:4}, 6) MOON \cite{f:5} and 7)  FedDyn \cite{f:8}. For a fair comparison, we share the same network, training details and data heterogeneity for all methods. 

\subsection{Comparison Experiments}
\subsubsection{Comparison Results}
As shown in Tables~\ref{tab: scanet},~\ref{tab: sunrgbd} and Fig.~\ref{fig: visual},  we present experimental results on SUN RGB-D and ScanNet V2 datasets to valuate superiority of our method under various settings of federated 3D object detection. The results under all settings suggest that Fed3D significantly outperforms existing FL methods $0.2\%\sim3.9\%$ mAP@0.25 and $0.1\%\sim3.0\%$ AR@0.25 on SUN RGB-D dataset. Furthermore, Fed3D achieves a better performance over compared methods about $0.8\%\sim4.8\%$ and $0.4\%\sim2.6\%$ in terms of mAP@0.25 and mAP@0.5 on ScanNet v2 datasets. As for communication cost, Fed3D requires 43.6M of communication, which is 50\% of other FL methods.  Such large improvement valuates  the superiority of our model to reduce communication cost. 

\subsubsection{Ablation Studies}
To demonstrate effectiveness of each module in Fed3D, Table~\ref{tab: ablation} shows ablation experiments under various ablation setting. w/o FPL,  LCC and GCC indicate the results of our model without federated prompt learning, local correction coefficient and global correction coefficient. Compared with Ours,  all
ablation variants severely degrade $0.3\%\sim4.8\%$ mAP@0.25. It verifies importance of the local-global class-aware correction module to address 3D data imbalance and the effectiveness of federated prompt learning to reduce communication cost.  

\subsection{Conclusion}
In this paper, we propose a first federated 3D object detection problem, and develop a Fed3D to effectively address two challenges (\emph{i.e.}, 3D data heterogeneity and limited communication
bandwidth). Specifically, we propose a local-global class-aware loss to tackle 3D data heterogeneity through perform balance local training from local and global aspects. Meanwhile, we develop a federated prompt learning to significantly reduce communication cost, which could transmitted a few parameters between global server and local clients. Comparison results demonstrate the superiority of our model to perform federated 3D object detection.

\newpage
\bibliographystyle{IEEEtran}
\bibliography{ref}{}

@article{f:1,
  title={Advances and open problems in federated learning},
  author={Kairouz, Peter and McMahan, H Brendan and Avent, Brendan and Bellet, Aur{\'e}lien and Bennis, Mehdi and Bhagoji, Arjun Nitin and Bonawitz, Kallista and Charles, Zachary and Cormode, Graham and Cummings, Rachel and others},
  journal={Foundations and Trends{\textregistered} in Machine Learning},
  volume={14},
  number={1--2},
  pages={1--210},
  year={2021},
  publisher={Now Publishers, Inc.}
}

@inproceedings{f:2,
  title={Communication-efficient learning of deep networks from decentralized data},
  author={McMahan, Brendan and Moore, Eider and Ramage, Daniel and Hampson, Seth and y Arcas, Blaise Aguera},
  booktitle={AISTATS},
  pages={1273--1282},
  year={2017},
  organization={PMLR}
}

@article{f:3,
  title={Federated optimization in heterogeneous networks},
  author={Li, Tian and Sahu, Anit Kumar and Zaheer, Manzil and Sanjabi, Maziar and Talwalkar, Ameet and Smith, Virginia},
  journal={Proceedings of Machine learning and systems},
  volume={2},
  pages={429--450},
  year={2020}
}

@inproceedings{f:4,
  title={Scaffold: Stochastic controlled averaging for federated learning},
  author={Karimireddy, Sai Praneeth and Kale, Satyen and Mohri, Mehryar and Reddi, Sashank and Stich, Sebastian and Suresh, Ananda Theertha},
  booktitle={ICML},
  pages={5132--5143},
  year={2020},
  organization={PMLR}
}

@inproceedings{f:5,
  title={Model-contrastive federated learning},
  author={Li, Qinbin and He, Bingsheng and Song, Dawn},
  booktitle={CVPR},
  pages={10713--10722},
  year={2021}
}

@inproceedings{f:6,
  title={Fedvision: An online visual object detection platform powered by federated learning},
  author={Liu, Yang and Huang, Anbu and Luo, Yun and Huang, He and Liu, Youzhi and Chen, Yuanyuan and Feng, Lican and Chen, Tianjian and Yu, Han and Yang, Qiang},
  booktitle={AAAI},
  volume={34},
  number={08},
  pages={13172--13179},
  year={2020}
}

@inproceedings{d:1,
  title={Deep hough voting for 3d object detection in point clouds},
  author={Qi, Charles R and Litany, Or and He, Kaiming and Guibas, Leonidas J},
  booktitle={ICCV},
  pages={9277--9286},
  year={2019}
}

@article{d:2,
  title={Pointnet++: Deep hierarchical feature learning on point sets in a metric space},
  author={Qi, Charles Ruizhongtai and Yi, Li and Su, Hao and Guibas, Leonidas J},
  journal={Advances in neural information processing systems},
  volume={30},
  year={2017}
}

@inproceedings{d:3,
  title={Mlcvnet: Multi-level context votenet for 3d object detection},
  author={Xie, Qian and Lai, Yu-Kun and Wu, Jing and Wang, Zhoutao and Zhang, Yiming and Xu, Kai and Wang, Jun},
  booktitle={CVPR},
  pages={10447--10456},
  year={2020}
}

@inproceedings{d:4,
  title={Voxelnet: End-to-end learning for point cloud based 3d object detection},
  author={Zhou, Yin and Tuzel, Oncel},
  booktitle={CVPR},
  pages={4490--4499},
  year={2018}
}

@inproceedings{d:5,
  title={An end-to-end transformer model for 3d object detection},
  author={Misra, Ishan and Girdhar, Rohit and Joulin, Armand},
  booktitle={ICCV},
  pages={2906--2917},
  year={2021}
}

@inproceedings{d:6,
  title={Group-free 3d object detection via transformers},
  author={Liu, Ze and Zhang, Zheng and Cao, Yue and Hu, Han and Tong, Xin},
  booktitle={ICCV},
  pages={2949--2958},
  year={2021}
}

@inproceedings{d:7,
  title={Polarformer: Multi-camera 3d object detection with polar transformer},
  author={Jiang, Yanqin and Zhang, Li and Miao, Zhenwei and Zhu, Xiatian and Gao, Jin and Hu, Weiming and Jiang, Yu-Gang},
  booktitle={AAAI},
  volume={37},
  number={1},
  pages={1042--1050},
  year={2023}
}

@inproceedings{d:8,
  title={3d object detection with pointformer},
  author={Pan, Xuran and Xia, Zhuofan and Song, Shiji and Li, Li Erran and Huang, Gao},
  booktitle={CVPR},
  pages={7463--7472},
  year={2021}
}

@inproceedings{d:9,
  title={Mvx-net: Multimodal voxelnet for 3d object detection},
  author={Sindagi, Vishwanath A and Zhou, Yin and Tuzel, Oncel},
  booktitle={ICRA},
  pages={7276--7282},
  year={2019},
  organization={IEEE}
}

@inproceedings{d:10,
  title={Swformer: Sparse window transformer for 3d object detection in point clouds},
  author={Sun, Pei and Tan, Mingxing and Wang, Weiyue and Liu, Chenxi and Xia, Fei and Leng, Zhaoqi and Anguelov, Dragomir},
  booktitle={ECCV},
  pages={426--442},
  year={2022},
  organization={Springer}
}

@inproceedings{d:11,
  title={Fcos3d: Fully convolutional one-stage monocular 3d object detection},
  author={Wang, Tai and Zhu, Xinge and Pang, Jiangmiao and Lin, Dahua},
  booktitle={ICCV},
  pages={913--922},
  year={2021}
}

@article{f:8,
  title={Federated learning based on dynamic regularization},
  author={Acar, Durmus Alp Emre and Zhao, Yue and Navarro, Ramon Matas and Mattina, Matthew and Whatmough, Paul N and Saligrama, Venkatesh},
  journal={arXiv preprint arXiv:2111.04263},
  year={2021}
}

@inproceedings{f:10,
  title={Ensemble attention distillation for privacy-preserving federated learning},
  author={Gong, Xuan and Sharma, Abhishek and Karanam, Srikrishna and Wu, Ziyan and Chen, Terrence and Doermann, David and Innanje, Arun},
  booktitle={ICCV},
  pages={15076--15086},
  year={2021}
}

@inproceedings{f:12,
  title={Privacy-preserving federated brain tumour segmentation},
  author={Li, Wenqi and Milletar{\`\i}, Fausto and Xu, Daguang and Rieke, Nicola and Hancox, Jonny and Zhu, Wentao and Baust, Maximilian and Cheng, Yan and Ourselin, S{\'e}bastien and Cardoso, M Jorge and others},
  booktitle={MLMI},
  pages={133--141},
  year={2019},
  organization={Springer}
}

@inproceedings{f:14,
  title={Fine-tuning global model via data-free knowledge distillation for non-iid federated learning},
  author={Zhang, Lin and Shen, Li and Ding, Liang and Tao, Dacheng and Duan, Ling-Yu},
  booktitle={CVPR},
  pages={10174--10183},
  year={2022}
}

@inproceedings{f:15,
  title={Fedproto: Federated prototype learning across heterogeneous clients},
  author={Tan, Yue and Long, Guodong and Liu, Lu and Zhou, Tianyi and Lu, Qinghua and Jiang, Jing and Zhang, Chengqi},
  booktitle={AAAI},
  volume={36},
  number={8},
  pages={8432--8440},
  year={2022}
}

@inproceedings{f:16,
  title={Local learning matters: Rethinking data heterogeneity in federated learning},
  author={Mendieta, Matias and Yang, Taojiannan and Wang, Pu and Lee, Minwoo and Ding, Zhengming and Chen, Chen},
  booktitle={CVPR},
  pages={8397--8406},
  year={2022}
}

@inproceedings{zhou2018voxelnet,
  title={Voxelnet: End-to-end learning for point cloud based 3d object detection},
  author={Zhou, Yin and Tuzel, Oncel},
  booktitle={CVPR},
  pages={4490--4499},
  year={2018}
}

@inproceedings{chen2017multi,
  title={Multi-view 3d object detection network for autonomous driving},
  author={Chen, Xiaozhi and Ma, Huimin and Wan, Ji and Li, Bo and Xia, Tian},
  booktitle={CVPR},
  pages={1907--1915},
  year={2017}
}

@article{cong2021comprehensive,
  title={A comprehensive study of 3-D vision-based robot manipulation},
  author={Cong, Yang and Chen, Ronghan and Ma, Bingtao and Liu, Hongsen and Hou, Dongdong and Yang, Chenguang},
  journal={IEEE Transactions on Cybernetics},
  volume={53},
  number={3},
  pages={1682--1698},
  year={2021},
  publisher={IEEE}
}

@inproceedings{rukhovich2022fcaf3d,
  title={Fcaf3d: Fully convolutional anchor-free 3d object detection},
  author={Rukhovich, Danila and Vorontsova, Anna and Konushin, Anton},
  booktitle={ECCV},
  pages={477--493},
  year={2022},
  organization={Springer}
}

@inproceedings{zhang2020h3dnet,
  title={H3dnet: 3d object detection using hybrid geometric primitives},
  author={Zhang, Zaiwei and Sun, Bo and Yang, Haitao and Huang, Qixing},
  booktitle={ECCV},
  pages={311--329},
  year={2020},
  organization={Springer}
}

@inproceedings{song2015sun,
  title={Sun rgb-d: A rgb-d scene understanding benchmark suite},
  author={Song, Shuran and Lichtenberg, Samuel P and Xiao, Jianxiong},
  booktitle={CVPR},
  pages={567--576},
  year={2015}
}

@inproceedings{dai2017scannet,
  title={Scannet: Richly-annotated 3d reconstructions of indoor scenes},
  author={Dai, Angela and Chang, Angel X and Savva, Manolis and Halber, Maciej and Funkhouser, Thomas and Nie{\ss}ner, Matthias},
  booktitle={CVPR},
  pages={5828--5839},
  year={2017}
}

@inproceedings{chen2023pimae,
  title={Pimae: Point cloud and image interactive masked autoencoders for 3d object detection},
  author={Chen, Anthony and Zhang, Kevin and Zhang, Renrui and Wang, Zihan and Lu, Yuheng and Guo, Yandong and Zhang, Shanghang},
  booktitle={CVPR},
  pages={5291--5301},
  year={2023}
}

@inproceedings{wang2021addressing,
  title={Addressing class imbalance in federated learning},
  author={Wang, Lixu and Xu, Shichao and Wang, Xiao and Zhu, Qi},
  booktitle={AAAI},
  volume={35},
  number={11},
  pages={10165--10173},
  year={2021}
}

@inproceedings{dong2022federated,
  title={Federated class-incremental learning},
  author={Dong, Jiahua and Wang, Lixu and Fang, Zhen and Sun, Gan and Xu, Shichao and Wang, Xiao and Zhu, Qi},
  booktitle={CVPR},
  pages={10164--10173},
  year={2022}
}

@article{li2021prefix,
  title={Prefix-tuning: Optimizing continuous prompts for generation},
  author={Li, Xiang Lisa and Liang, Percy},
  journal={arXiv preprint arXiv:2101.00190},
  year={2021}
}

@inproceedings{thapa2022splitfed,
  title={Splitfed: When federated learning meets split learning},
  author={Thapa, Chandra and Arachchige, Pathum Chamikara Mahawaga and Camtepe, Seyit and Sun, Lichao},
  booktitle={AAAI},
  volume={36},
  number={8},
  pages={8485--8493},
  year={2022}
}

@inproceedings{liu2021feddg,
  title={Feddg: Federated domain generalization on medical image segmentation via episodic learning in continuous frequency space},
  author={Liu, Quande and Chen, Cheng and Qin, Jing and Dou, Qi and Heng, Pheng-Ann},
  booktitle={CVPR},
  pages={1013--1023},
  year={2021}
}

@article{yang2021federated,
  title={Federated semi-supervised learning for COVID region segmentation in chest CT using multi-national data from China, Italy, Japan},
  author={Yang, Dong and Xu, Ziyue and Li, Wenqi and Myronenko, Andriy and Roth, Holger R and Harmon, Stephanie and Xu, Sheng and Turkbey, Baris and Turkbey, Evrim and Wang, Xiaosong and others},
  journal={Medical image analysis},
  volume={70},
  pages={101992},
  year={2021},
  publisher={Elsevier}
}

@ARTICLE{sun2022data,
  author={Sun, Gan and Cong, Yang and Dong, Jiahua and Wang, Qiang and Lyu, Lingjuan and Liu, Ji},
  journal={IEEE Internet of Things Journal}, 
  title={Data Poisoning Attacks on Federated Machine Learning}, 
  year={2022},
  volume={9},
  number={13},
  pages={11365-11375},
  doi={10.1109/JIOT.2021.3128646}}
\end{document}